\documentclass[10pt,letterpaper,conference]{ieeeconf}
\usepackage[latin9]{inputenc}
\usepackage{array}
\usepackage{textcomp}
\usepackage{multirow}
\usepackage{amsmath}
\usepackage{graphicx}
\let\labelindent\relax
\usepackage{enumitem}
\usepackage{cite}
\usepackage{hyperref}

\usepackage{eso-pic}

\makeatletter

\providecommand{\tabularnewline}{\\}

\IEEEoverridecommandlockouts                              
\hypersetup{hidelinks}

\makeatother

\begin{document}

\thispagestyle{empty}\pagestyle{empty} 

\title{\textbf{Evaluating the Impact of Semantic Segmentation and Pose Estimation on Dense Semantic SLAM }}

\author{Suman Raj Bista, David Hall, Ben Talbot, Haoyang Zhang,
Feras Dayoub and Niko S\"{u}nderhauf\thanks{The authors are with Queensland University of Technology (QUT). This research has been conducted by the  Australian Research Council Centre of Excellence for Robotic Vision (CE140100016) and supported by the QUT Centre for Robotics. \texttt{suman.bista@qut.edu.au}}}

\AddToShipoutPicture*{%
     \AtTextUpperLeft{%
         \put(-3.5,10){
           \begin{minipage}{\textwidth}
              \footnotesize \textbf{\\This paper has been accepted for presentation at the 2021 IEEE/RSJ International Conference on Intelligent Robots and Systems (IROS 2021) in Prague, Czech Republic, Online (September 27- October 1, 2021).\\
              Final version available at \url{http://ieeexplore.ieee.org/}}
           \end{minipage}}%
     }%
}

\maketitle

\begin{abstract}
Recent Semantic SLAM methods combine classical geometry-based estimation with deep learning-based object detection or semantic segmentation.
In this paper we evaluate the quality of semantic maps generated by state-of-the-art class- and instance-aware dense semantic SLAM algorithms whose codes are publicly available and explore the impacts both semantic segmentation and pose estimation have on the quality of semantic maps.
We obtain these results by providing algorithms with ground-truth pose and/or semantic segmentation data available from simulated environments. We establish that semantic segmentation is the largest source of error through our experiments, dropping mAP and OMQ performance by up to 74.3\% and 71.3\% respectively.
\end{abstract}

\section{INTRODUCTION}

Semantic scene understanding is an important capability for robotic systems.
The ability to determine \textit{what} is \textit{where} within an environment is a fundamental building block of operation.
A chief problem within this field of research is semantic simultaneous localization and mapping (SLAM) and, within that, the semantic mapping of objects in 3D space.
Utilizing the power of deep learning, there have recently been several influential works in this area capable of providing dense semantic maps~\cite{salas2013slam++, Rosinol20icra-Kimera, mccormac2017semanticfusion, maskfusion,grinvald2019volumetric,hachiuma2019detectfusion,narita2019panopticfusion}.
However, there is limited work in examining the performance of such systems, particularly in understanding the effects of internal components on semantic mapping performance and quality.

In this work, we examine class-aware and instance-aware dense semantic SLAM algorithms to determine what components cause the greatest errors in creating a semantic map.
Specifically, we examine the effect of class/instance segmentation and localization on the performance of generating semantic maps.
This is a level of analysis for semantic mapping that is rarely seen in the literature.

We are able to analyze sub-components of dense semantic SLAM systems to this level of detail in isolation by using simulated data.
While some standard real-world datasets exist such as in~\cite{geiger2012cvpr,behley2019iccv}, we leverage advances in simulated environments for robotics tasks (like ~\cite{ai2thor,habitat19iccv}) to attain repeatable results in environments with controlled variations.
We use the BenchBot framework~\cite{talbot2020benchbot} to gain access to ground-truth instance segmentation, depth, and pose data within a set of simulated environments~\cite{hall2020robotic}.
By selectively utilizing this ground-truth data within the semantic mapping pipeline, we are able to examine which components have the highest impact on semantic mapping and compare the different mapping techniques.

The key findings of this paper are as follows:
\begin{itemize}
    \item Semantic segmentation/object segmentation is the main cause of error in semantic mapping systems;
    \item Class-aware SLAM systems attain better results than instance-aware systems under ideal conditions due to better data association; and
    \item Providing perfect segmentation, depth and pose data does not guarantee perfect semantic mapping.
\end{itemize}

The following section gives an overview of some semantic mapping methods
and their evaluation metrics available in the literature. Our approach
for the evaluation, along with the methods and evaluation metrics
we have used, are described in Section III. Section IV presents and discusses our
experimental results. Finally, some
concluding remarks are provided in Section V.

\section{Background Literature}
As our work concerns the evaluation of semantic mapping systems, we provide background research in both semantic SLAM/semantic mapping systems and the metrics that are used to evaluate them.

\subsection{Semantic SLAM / Semantic Mapping Systems}
Dense SLAM with semantic labels and instance-aware SLAM systems have been developed in recent years by utilizing developments in deep learning, object detection and
semantic segmentation. SLAM++ \cite{salas2013slam++}, is one of the first RGB-D object-oriented mapping system. 
Although a pioneering work, SLAM++ requires a database of well-defined 3D models of the objects (full set of object instances with their very detailed geometric shapes) beforehand, which is overly restrictive for widespread application. 
Comparatively, SemanticFusion \cite{mccormac2017semanticfusion} builds a dense semantic map by fusing semantic predictions from a convolutional neural network into a dense map built using RGB-D SLAM (ElasticFusion \cite{whelan2016elasticfusion}) without needing prior 3D models. 
Recently \cite{Rosinol20icra-Kimera} built a dense volumetric
semantic map incorporating semantic segmentation in the Voxblox framework
proposed by \cite{oleynikova2017voxblox}.
Both methods are class-aware dense semantic SLAM systems that focus on the general class label given to each pixel in an image rather than individual objects within a scene.

As well as class-aware semantic SLAM systems, some approaches attempt to create semantic maps from instance-aware object detections.
MaskFusion \cite{maskfusion} builds an instance-aware map by fusing semantic labels from instance-aware semantic segmentation into the RGB-D SLAM system. 
Fusion++ \cite{mccormac2018fusion++} uses Mask R-CNN \cite{he2017mask} instance segmentation to initialize compact per-object Truncated Signed Distance Function (TSDF) reconstructions to build a semantic graph-based map. 
The consistency of the map is then maintained via loop closure detection, pose-graph optimization and further refinement. 
An instance-aware semantic map was also developed in~\cite{grinvald2019volumetric} by using instance-aware segmentation within RGB images and unsupervised depth segmentation. 
The work in~\cite{hachiuma2019detectfusion} uses both 2D object detection and 3D geometric segmentation to build an object map.
PanopticFusion \cite{narita2019panopticfusion} goes further than just using either class- or instance-aware segmentation, instead utilizing both to build a 3D pantopic map by fusing 2D semantic and instance segmentation
outputs refined using dense Conditional Random Field (CRF). 

There is also a growing field of works on instance-aware SLAM with sparse representation of the object using geometric primitives like cuboids \cite{yang2019cubeslam} and quadrics \cite{nicholson2018quadricslam}. 
Both CubeSLAM \cite{yang2019cubeslam} and QuadricSLAM~\cite{nicholson2018quadricslam} are monocular object SLAM systems, which rely on the bounding boxes generated by object detection algorithms to estimate the initial cuboid proposal and the quadric parameters respectively. 
These sparse SLAM systems improve localization accuracy but have their own limitations. 
In CubeSLAM, vanishing points are also required during cuboid proposal generation, limiting the applicable object categories. 
The extraction of vanishing points requires the objects to have parallel straight lines on the surface. 
QuadricSLAM requires the same object to be seen from at least three views to initialize quadric parameters. 
Furthermore, diverse viewpoints of the objects are required for convergence. The straight-line motions of the robot may not always generate diverse enough viewpoints of objects. Hence, we do not consider these sparse SLAM systems in our evaluation.

\subsection{Evaluation Metrics}
Visual SLAM methods are typically evaluated using
Absolute Trajectory Error (ATE) and the Relative Pose Error (RPE)~\cite{sturm2012benchmark}.
ATE evaluates the global consistency of trajectories by comparing positional
offsets between the estimated and the ground-truth trajectories. 
In contrast, RPE considers the drift of trajectory or local motion errors in
the trajectory. Both ATE and RPE are usually expressed as Root-Mean-Square-Error (RMSE) \cite{sturm2012benchmark,mur2017orb} and are beneficial in testing localization accuracy. 
However, these metrics do not evaluate the semantics or quality of object maps. 

With the development of more semantic mapping systems, some approaches have been put forward to evaluate the quality of SLAM maps and the trajectories they produce.
SemanticFusion \cite{mccormac2017semanticfusion} uses pixel average accuracy (percentage of correctly classified pixels) and class average accuracy (the average of the diagonal of the prediction\textquoteright s
normalized confusion matrix). 
MaskFusion \cite{maskfusion} uses 3D error for reconstruction evaluation and Intersection-over-Union (IoU)
\cite{everingham2010pascal} using the reprojected object mask of the
reconstructed 3D model for segmentation evaluation. 
PanopticFusion \cite{narita2019panopticfusion} uses panoptic quality, segmentation quality and recognition quality calculated based on IoU over 3D segmentation masks.
In QuadricSLAM \cite{nicholson2018quadricslam}, centroid error is used to measure position quality, and the Jaccard distance (1\textminus IoU) is used to measure shape quality. 
The metrics such as accuracy of centroid estimation and IoU evaluate only the spatial quality, whereas semantic maps also have to be evaluated semantically.

Inspired by 2D object detection, the fusion of semantic and spatial
analysis is typically done using adaptations of mean average precision
(mAP) with IoU in 3D used in place of 2D IoU \cite{yang2019cubeslam,hou20193d,qi2019deep,yang20203dssd}. In \cite{grinvald2019volumetric}, Average Precision (AP) \cite{everingham2010pascal} is computed for each class using an IoU threshold of 0.5 over the predicted 3D segmentation masks. 
CubeSLAM \cite{yang2019cubeslam} uses 3D IoU and AP calculated using an IoU threshold of 0.25 for evaluating estimated cuboids. 
Recently, \cite{hall2020robotic} extends the Probability-based Detection
Quality (PDQ) \cite{hall2020probabilistic} evaluation measure designed
for probabilistic object detection to evaluate object-based semantic
map. This Object Map Quality (OMQ) evaluation method is used as the
metrics in the Scene Understanding Challenge \cite{hall2020robotic}.
Since mAP using 3D IoU and OMQ can both evaluate spatial and semantic
quality with a single metric, we choose to use them in evaluating the performance of some of the popular semantic mapping algorithms in this work.

\section{METHOD OVERVIEW}
To establish the impact of semantic segmentation and pose estimation on different semantic SLAM systems, we perform experimental analysis in a set of controlled simulated environments.
All experiments were performed on a PC with Intel Core i7 CPU and 16 GB RAM equipped with Nvidia GeForce GTX 1080 GPU having 8 GB graphics memory.

\subsection{Simulation Environments}
\begin{figure}[t]
\includegraphics[width=1\columnwidth]{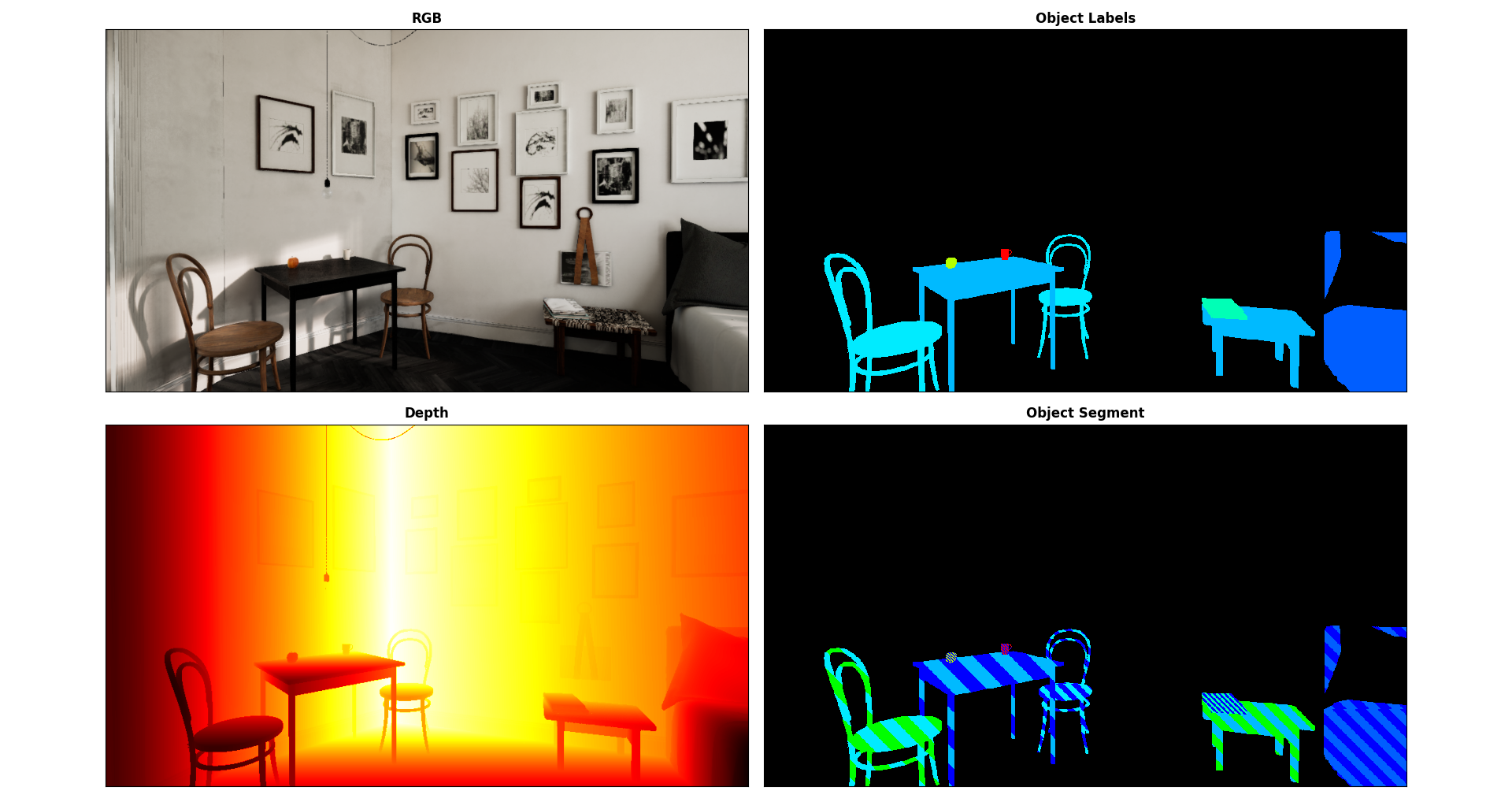}\caption{Image output from the BenchBot simulator: RGB (top-left), Depth (bottom-left), Class Segmentation (top-right), Instance Segmentation (bottom-right). Note that instance segmentation visualization has stripes depicting class and instance id of a given object, but actual output is a single instance value without stripes.}
\label{fig:sim_output}
\vspace{-5mm}
\end{figure}

In this paper, we used the high-fidelity simulator in the BenchBot \cite{talbot2020benchbot}
system to obtain color and depth images, ground-truth
poses, and ground-truth class and object segmentation. Example outputs are shown in Fig.~\ref{fig:sim_output}. 
The BenchBot system provides access to five simulated indoor environments called miniroom, apartment, house, office and company. Each environment has five variations: base, day, night, day extra and night extra. The variations present changes in both lightning and the objects present within each scene. 
This system has been primarily designed for the Scene Understanding Challenge and provides ground-truth 3D cuboid maps for 30 classes of object : bottle, cup, knife, bowl, wine glass, fork, spoon, banana, apple, orange, cake, potted plant, mouse, keyboard, laptop, cellphone, book, clock, chair, dining table, couch, bed, toilet, monitor, microwave, toaster, refrigerator, oven, sink, and person \cite{hall2020robotic}. However, we were unable to use all of the objects in the larger environments due to GPU memory requirements of some of the methods being evaluated. As a result, we had to choose a subset of the available simulated environments.
We utilize the miniroom and apartment environments as they are small and medium-sized environments respectively.
We perform experiments on one day scene (1:base) and one night scene (3:night) for both of these environments.
Altogether, the algorithms are evaluated in the apartment:1, apartment:3, miniroom:1 and miniroom:3 scenes within BenchBot. 

\subsection{Investigated SLAM Systems}
Our work examines some of the popular open-source Semantic Mapping/SLAM methods
that have publicly available code. 
We have used the pre-trained models in standard datasets available publicly
for object detection and semantic segmentation. 
Of the available dense class-aware semantic SLAM methods, we examine SemanticFusion (SF)\cite{mccormac2017semanticfusion} and KimeraSemantics (KS)\cite{Rosinol20icra-Kimera}. 
For semantic segmentation, we use DilatedNet pertained in the ADE20K dataset \cite{zhou2016semantic}.
We have specifically chosen this segmentation because of its pre-trained
model available in Caffe \cite{jia2014caffe} framework used by SemanticFusion \cite{mccormac2017semanticfusion}. 
The instance-aware semantic SLAM approaches analysed in this work are MaskFusion (MF)~\cite{maskfusion} and Voxblox++ (Vpp)~\cite{grinvald2019volumetric}.
Fusion++ \cite{mccormac2018fusion++} was not used due to an absence of publicly available code. 
For these, the instance-level segmentation is provided by
Mask R-CNN \cite{he2017mask} pre-trained on the MS COCO dataset \cite{lin2014microsoft}.
For all methods, the estimated poses are obtained from ElasticFusion~\cite{whelan2016elasticfusion}
when ground-truth poses are not used. 

\subsubsection{SemanticFusion (SF)\cite{mccormac2017semanticfusion}}

SemanticFusion combines semantic predictions from deep-learned networks
with ElasticFusion \cite{whelan2016elasticfusion}. ElasticFusion is a real time RGB-D SLAM that provides dense correspondences between frames and a globally consistent map of fused surfels. The semantic predictions consist of a set of per-pixel class probabilities.
The class probability distribution for each surfel is updated using
Bayesian update from the semantic predictions and correspondences
provided by the SLAM system. The semantic predictions in the map are further
improved using fully-connected CRF regularization. We have used publicly available
SemanticFusion code \cite{sfusion}, which uses the Caffe
framework \cite{jia2014caffe} for semantic predictions. The original
code uses a pre-trained model in the NYUv2 dataset, consisting of
only 12 classes that do not cover all the classes required for our tests. Consequently, we used the semantic predictions from DilatedNet, which is pre-trained on
the ADE20K dataset \cite{zhou2016semantic} and also available in Caffe.
To make \cite{zhou2016semantic} compatible with SemanticFusion, we added a softmax layer after the final up-sampled fully connected
layer, which outputs per-pixel class probabilities.  When using ground-truth
segmentation from Benchbot, we generate per-pixel class probability
from ground-truth masks where the class probability for the ground-truth
class is assigned as 1.0.

\subsubsection{MaskFusion (MF)\cite{maskfusion}}

MaskFusion uses the object masks obtained from instance-level semantic segmentation to create an instance-aware semantic map, with each object in the 3D map represented as a set of surfels. The object mask
is derived by combining the geometric segmentation with the instance-level
semantic segmentation. The geometric segmentation is obtained from
the analysis of depth discontinuities and surface normals, whereas
the instance-level semantic segmentation is obtained either from Mask R-CNN
\cite{he2017mask} (pre-trained on the MS COCO dataset \cite{lin2014microsoft})
or from BenchBot's ground-truth segmentation. We used the code of MaskFusion available in \cite{mfusion}, which
requires a high-end GPU with enough memory to store multiple models
simultaneously for smooth real-time performance. Therefore,
the GPU memory limited the size of the environment and the number of the
objects that could be mapped successfully. 

\subsubsection{Voxblox++ (Vpp)\cite{grinvald2019volumetric}}

Voxblox++ creates volumetric object-centric maps incrementally. The frame-wise instance segmentation is obtained by
combining the unsupervised geometric segmentation of depth images
with supervised semantic object predictions from RGB. The data association
strategy tracks the individual predicted instances across multiple
frames. Finally, observed surface geometry and semantic information
are fused into a global TSDF map volume. We have used the code of
Voxblox++ available in \cite{vpp}, where \cite{depth_segmentation}
is used for depth segmentation and either Mask R-CNN (pre-trained
in the MS COCO dataset) or BenchBot's ground-truth segmentation
is used for instance segmentation. Except
for the Mask R-CNN component, all other stages of Voxblox++ can run online on a
CPU.

\subsubsection{KimeraSemantics (KS)\cite{Rosinol20icra-Kimera}}

KimeraSemantics uses semantically labeled images to produce a lightweight metric-semantic mesh model
of the environment. The semantic
labels obtained from 2D semantic segmentation are attached to each
point in the 3D point cloud, which is obtained from RGB-D or dense stereo. Next,
bundled raycasting from Voxblox \cite{oleynikova2017voxblox} is
applied, with the label probabilities built from the frequency
of the observed labels in each bundle. Bayesian updates are then used
to update the label probabilities at each voxel while traversing the
voxels along the ray. TSDF is used to filter out noise and extract
the global mesh. We used the code of KimeraSemantics available
in \cite{ksemantics}, where 2D semantic segmentation is obtained
from DilatedNet (pre-trained on the ADE20K dataset)
or BenchBot's ground-truth segmentation. This method
is modular and can run in real-time on a CPU.

To evaluate against the 3D object cuboid ground-truth maps supplied by BenchBot, we require each tested method to output instance-wise, semantically labeled, axis-aligned object cuboids.
The global maps obtained from SemanticFusion and
KimeraSemantics give 3D semantic point clouds at the class
level rather than the instance level. 
The instance point cloud for each class is extracted by segmenting the point cloud of the corresponding class based on the Euclidean distance using MATLAB's \texttt{pcsegdist} function \cite{pcsegdist}. 
MaskFusion and Voxblox++ already give point clouds of the object instances in the global map frame. 
The point cloud of each object instance is used to obtain axis-aligned cuboids in the world-coordinate frame that can be used within the evaluation.


\subsection{Evaluation Metrics\label{subsec:Evaluation-Metrics}}
\begin{figure}[h]
\vspace{-2mm}
\begin{centering}
\includegraphics[scale=0.5]{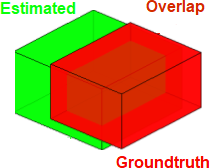}
\par\end{centering}
\caption{Visualization of calculating IoU in 3D. \label{fig:Calculation-of-IoU}}
\vspace{-2mm}
\end{figure}

Our work focuses on evaluating the semantic maps produced by semantic SLAM systems by comparing estimated object cuboids with their ground-truth counterparts.
To perform our analysis, we use two metrics: 
\begin{enumerate}[label=\arabic*)]
    \item mean Average Precision (mAP) \cite{everingham2010pascal,lin2014microsoft} calculated based on the IoU of cuboids, and 
    \item Object Map Quality (OMQ) \cite{hall2020robotic}. 
\end{enumerate}
Calculating the 3D IoU between ground-truth cuboids and those produced by the Semantic SLAM systems is a core component to both measures.
Given ground-truth cuboids and estimated cuboids axis-aligned in world coordinate frame parameterized by a centroid $(x,y,z)$ and the cuboids full extent $(l,b,w)$, we can compute IoU in 3D as follows: 

\begin{equation}
IoU_{3D}=\frac{V_{overlap}}{V_{gt}+V_{est}-V_{overlap}},
\end{equation}

\noindent where $V_{gt}$ is the volume of the ground-truth cuboid, $V_{est}$ is
the volume of the estimated cuboid, $V_{overlap}$ is the overlapped
volume as shown in Fig. \ref{fig:Calculation-of-IoU}.

Each of the evaluation metrics then has its own set of implementation details, which are outlined in the following two paragraphs:

\subsubsection{mAP using 3D IoU}
We calculate mAP using 3D IoU for IoU thresholds $0.25$ to $0.95$
at interval of $0.05$ (i.e. $0.25:0.05:0.95)$ similar to \cite{cocoeval}.
Mapped cuboids are considered true positives (TP) if: a) the IoU of the estimated and ground-truth cuboid exceeds the threshold, and b) both estimated and ground-truth objects share the same class. 
To find the closest match in the ground-truth list, we sort all the detections belonging to the same class in descending order of the confidence score and
search for the ground truth object in the same class with the highest
IoU. 
Precision-recall (PR) curves\cite{everingham2010pascal} for each class
are calculated at each of these $15$ IoU thresholds. A PR curve is then created using the average precision swept across all classes and IoU thresholds. 
Lastly, the final mAP score for the scene is given as the average of all of these precision values. We denote it by $\text{mAP}_{\text{\small3D}}$.
mAP is also calculated for the thresholds $0.25$ and $0.5$
for the scene as is done in~\cite{cocoeval}, and denoted by $\text{mAP}_{\text{\small3D}}^{\text{\small25}}$
and $\text{mAP}_{\text{\small3D}}^{\text{\small50}}$ respectively.

\subsubsection{OMQ}

Adapted from the probability-based detection quality (PDQ) evaluation
measure \cite{hall2020probabilistic}, OMQ \cite{hall2020robotic}
is a new measure for evaluating 3D object maps. A pairwise object quality (pOQ) is calculated
between each object in the proposed map and each ground-truth object in the ground-truth map.
The pOQ score is the geometric mean of spatial quality ($\mbox{Q}_{Sp}$)
and label quality ($\mbox{Q}_{L}$). For the $\mbox{i}^{th}$ proposed
object ($\mbox{O}_{i})$ and $\mbox{j}^{th}$ ground-truth object
($\hat{\mbox{O}_{j}}$), pOQ is defined as \vspace{-5mm}

\begin{equation}
pOQ(O_{i},\hat{O_{j}})=\sqrt{Q_{Sp}(O_{i},\hat{O_{j}}).Q_{L}(O_{i},\hat{O_{j}})}\,.
\end{equation}
$\mbox{Q}_{Sp}$ is simply the IoU score in 3D between ground-truth
and proposed object cuboids. $\mbox{Q}_{L}$ is the probability given
to the correct class. Once all pairwise scores are calculated, objects
in the proposed map are optimally assigned to objects in the ground-truth
map. From this, a list of true positive quality scores with non-zero
quality assignments ($\mbox{\textbf{q}}_{TP}$), and the number of
true positives ($\mbox{n}_{TP}$), false negatives ($\mbox{n}_{FN}$),
and false positives ($\mbox{n}_{FP}$) are obtained. In order to weigh
overconfident false positives as worse than low-confidence false positives,
a list of false positive costs ($\mbox{\textbf{c}}_{FP}$) for all
false positives is calculated. False positive cost is simply the maximum
label probability given to a non-background class label. The final
OMQ score is defined as

\begin{equation}
OMQ=\frac{{\displaystyle {\displaystyle {\textstyle \sum_{i=1}^{n_{TP}}\boldsymbol{q}_{TP}}(i)}}}{n_{TP}+n_{FN}+{\textstyle \sum_{i=1}^{n_{FP}}\boldsymbol{c}_{FP}}(i)}.
\end{equation}

\subsection{Testing Scenarios \label{subsec:Experimental-Design}}

For evaluation, we compared the estimated cuboid map generated
by the methods mentioned earlier against the ground-truth cuboid
map provided by the BenchBot in the following scenarios:
\begin{enumerate}
\item \textit{Case I}: Using all ground-truth data obtained from \cite{talbot2020benchbot},
i.e., ground-truth poses and ground-truth segmentation or ground-truth
object masks. This case is the ideal scenario that uses perfect localization
and detection that serves as a baseline for the given method.
\item \textit{Case II}: Using ground-truth poses and estimated class-based
\cite{zhou2016semantic} or object-based \cite{he2017mask} segmentation using
deep neural networks. This case is used to evaluate mapping performance
in the simulated environment using the network pre-trained on real-world data within a perfect localization scenario.
\item \textit{Case III}: Using estimated pose from dense RGB-D SLAM \cite{whelan2016elasticfusion}
and ground-truth segmentation or ground-truth object masks. This case
is used to evaluate how much the map quality is affected using the
estimated pose when detections are perfect.
\item \textit{Case IV}: Using estimated pose \cite{whelan2016elasticfusion}
and estimated segmentation \cite{zhou2016semantic} or object masks
\cite{he2017mask}. This case is used to evaluate how much the map
quality is affected in the realistic robotic scenario using estimated
poses and detections. 
\end{enumerate}

\section{EXPERIMENTS AND RESULTS\label{sec:EXPERIMENTS-AND-RESULTS}}

Tables \ref{tab:Using-all-ground-truth}, \ref{tab:Using-ground-truth-pose},
\ref{tab:Using-ground-truth-segmentation} and \ref{tab:Using-estimated-pose}
present the evaluation scores using metrics outlined in Sect. \ref{subsec:Evaluation-Metrics}
for Cases \textit{I}, \textit{II}, \textit{III} and \textit{IV} respectively.
The considered methods SemanticFusion, MaskFusion, Voxblox++ and KimeraSemantics
are denoted by SF, MF, Vpp and KS respectively. We denote environments
apartment:1, apartment:3, miniroom:1 and miniroom:3 as ap1, ap3, mr1
and mr3 respectively. The root mean square trajectory error between
the estimated pose obtained from \cite{whelan2016elasticfusion} and
the ground-truth pose are presented in Table \ref{tab:RMSE-ATE-and}.
We use Absolute Trajectory Error (ATE) and Relative Pose Error
(RPE) to measure the trajectory error. 

The metrics in Tables \ref{tab:Using-all-ground-truth}, \ref{tab:Using-ground-truth-pose}, \ref{tab:Using-ground-truth-segmentation} and \ref{tab:Using-estimated-pose} 
are as follows:\textbf{ }$\text{mAP}_{\text{\small3D}}$ is the average
mAP using 3D IoU for IoU threshold $0.25$ to $0.95$ at the interval
of $0.05$, $\text{mAP}_{\text{\small3D}}^{\text{\small25}}$ is the
average mAP using 3D IoU threshold of 0.25, $\text{mAP}_{\text{\small3D}}^{\text{\small50}}$
is the average mAP using 3D IoU threshold of 0.5, $\mbox{OMQ}$ is
Object Map Quality score,\textbf{ }mPOQ\textbf{ }is the average pairwise
quality of true positives, mLQ\textbf{ }is the average label quality
of true positives, and mFPQ\textbf{ }is the average spatial quality
(3D IoU) of true positives. The metrics rmAP, rAP25, rAP50, and rOMQ
are the ratio of $\text{mAP}_{\text{\small3D}}$, $\text{mAP}_{\text{\small3D}}^{\text{\small25}}$,
$\text{mAP}_{\text{\small3D}}^{\text{\small50}}$, and $\mbox{OMQ}$
respectively with their corresponding score calculated using all ground-truths.

\begin{figure}[h]
\vspace{-1mm}
\includegraphics[width=1\columnwidth]{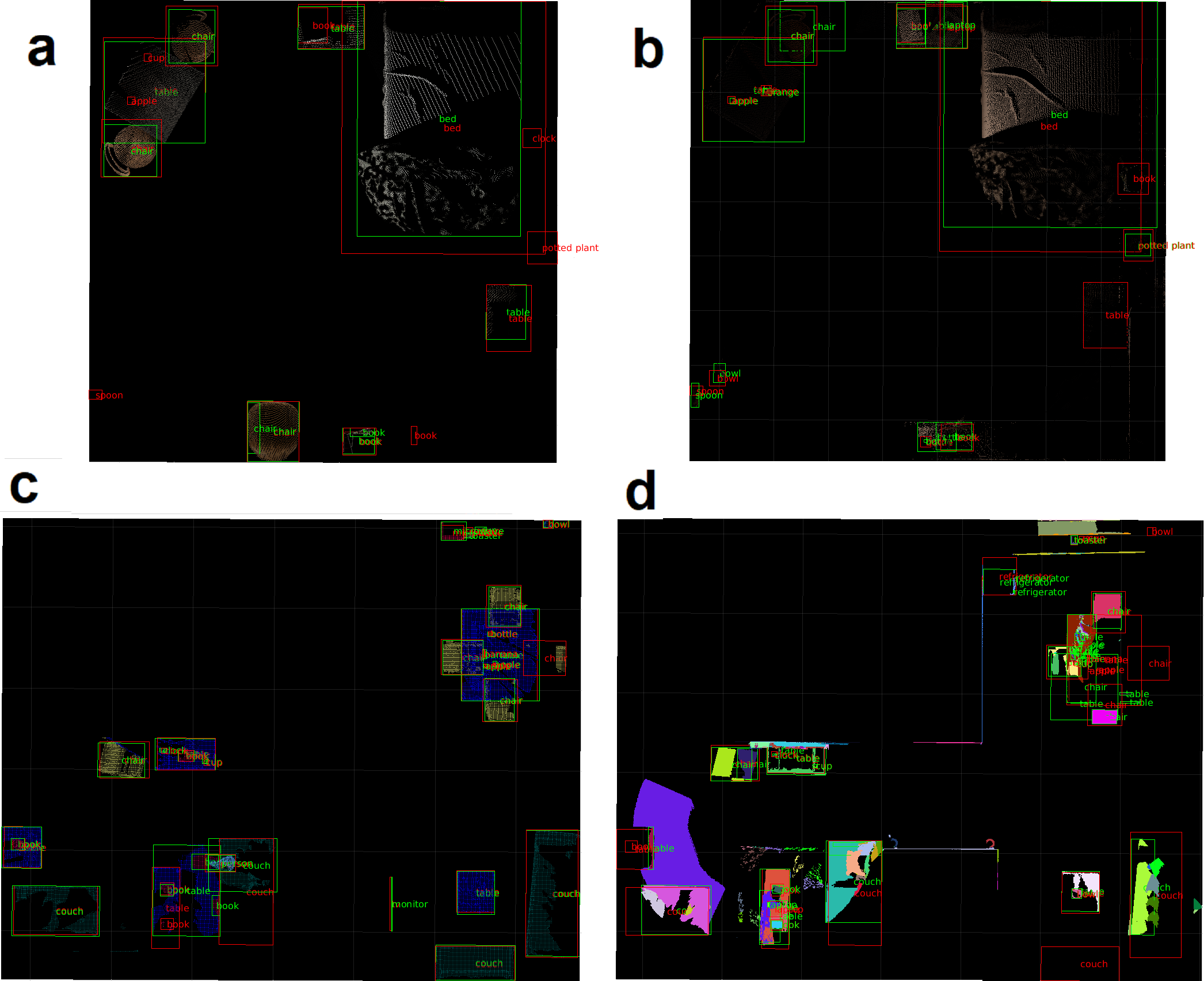}

\caption{Object Maps using all ground-truths. \textbf{a} MaskFusion on miniroom:1, \textbf{b}
SemanticFusion on miniroom:3, \textbf{c} KimeraSemantics on apartment:1, and
\textbf{d} Voxblox++ on apartment:3. Ground-truth cuboids are shown in red.
Estimated Cuboids are shown in green. \label{fig:Object-Maps-}}
\vspace{-2.5mm}
\end{figure}

\begin{figure}[t]
\includegraphics[width=1\columnwidth]{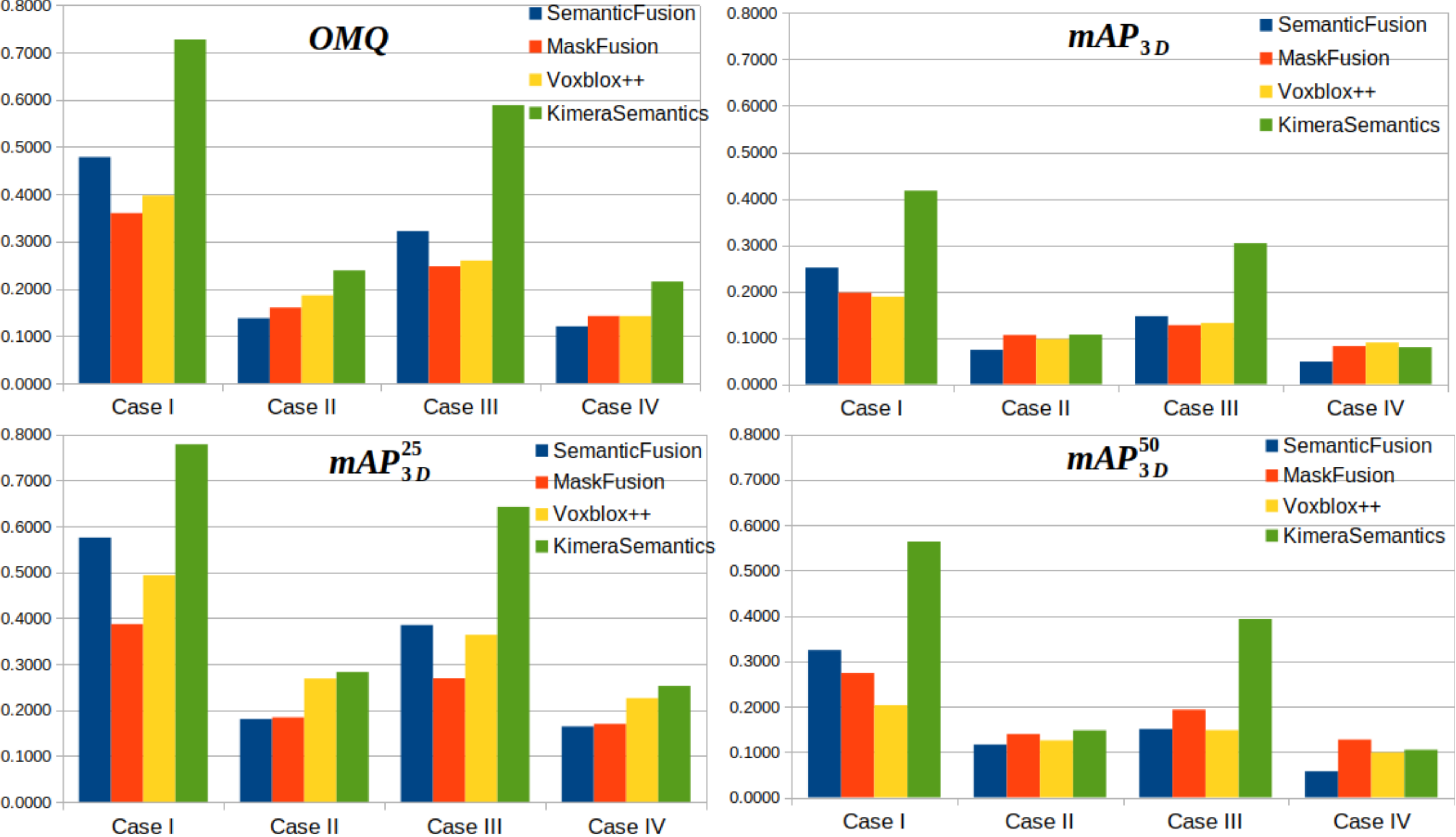}

\caption{OMQ and mAP scores based on IoU in 3D.\label{fig:scores}}

\vspace{-2mm}
\end{figure}
\begin{figure}
\includegraphics[width=1\columnwidth]{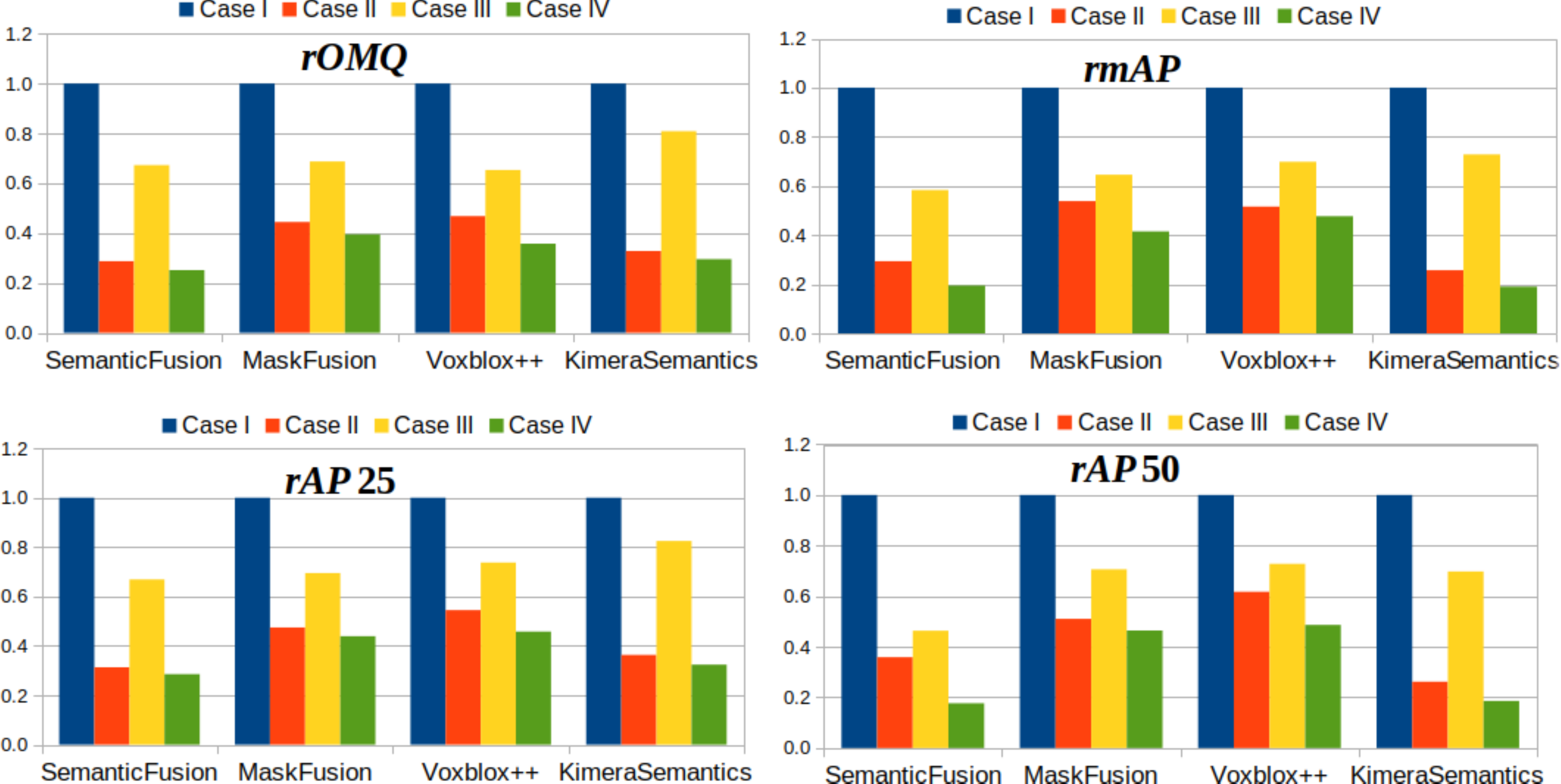}

\caption{Ratio of OMQ and mAP with their corresponding score calculated for
\textit{Case I}. \label{fig:rscore}}

\vspace{-5mm}
\end{figure}

\begin{table*}
\begin{centering}
\begin{tabular}{|c|c|c|c|c|c|c|c|c|c|}
\hline 
\multirow{2}{*}{\textbf{Method}} & \multirow{2}{*}{\textbf{Scene}} & \multirow{2}{*}{\textbf{$\mbox{\textbf{mAP}}_{\boldsymbol{\mbox{3D}}}$}} & \multirow{2}{*}{$\mbox{\textbf{mAP}}_{\boldsymbol{\mbox{3D}}}^{\mbox{\textbf{25}}}$} & \multirow{2}{*}{$\mbox{\textbf{mAP}}_{\boldsymbol{\mbox{3D}}}^{\mbox{\textbf{50}}}$} & \multirow{2}{*}{$\mbox{\textbf{OMQ}}$} & \multirow{2}{*}{\textbf{$\boldsymbol{\mbox{mPOQ}}$}} & \multirow{2}{*}{\textbf{$\boldsymbol{\mbox{mLQ}}$}} & \multirow{2}{*}{\textbf{$\boldsymbol{\mbox{mSQ}}$}} & \multirow{2}{*}{\textbf{$\boldsymbol{\mbox{mFPQ}}$}}\tabularnewline
 &  &  &  &  &  &  &  &  & \tabularnewline
\hline 
Semantic Fusion (SF) & \multirow{4}{*}{apartment:1 (ap1)} & 0.2641 & 0.5538 & 0.3772 & 0.5450 & 0.6812 & 1.0 & 0.4882 & 1.00\tabularnewline
\cline{1-1} \cline{3-10} \cline{4-10} \cline{5-10} \cline{6-10} \cline{7-10} \cline{8-10} \cline{9-10} \cline{10-10} 
Mask Fusion (MF) &  & 0.2004 & 0.2671 & 0.2671 & 0.3993 & 0.8486 & 1.0 & 0.7273 & 0.00\tabularnewline
\cline{1-1} \cline{3-10} \cline{4-10} \cline{5-10} \cline{6-10} \cline{7-10} \cline{8-10} \cline{9-10} \cline{10-10} 
Voxblox++ (Vpp) &  & 0.1268 & 0.3848 & 0.0830 & 0.5108 & 0.6385 & 1.0 & 0.4502 & 1.00\tabularnewline
\cline{1-1} \cline{3-10} \cline{4-10} \cline{5-10} \cline{6-10} \cline{7-10} \cline{8-10} \cline{9-10} \cline{10-10} 
Kimera Semantics (KS) &  & 0.3804 & 0.7402 & 0.4947 & 0.7473 & 0.7473 & 1.0 & 0.5942 & 1.00\tabularnewline
\hline 
Semantic Fusion (SF) & \multirow{4}{*}{apartment:3 (ap3)} & 0.1406 & 0.3986 & 0.1944 & 0.3568 & 0.5946 & 1.0 & 0.4062 & 0.00\tabularnewline
\cline{1-1} \cline{3-10} \cline{4-10} \cline{5-10} \cline{6-10} \cline{7-10} \cline{8-10} \cline{9-10} \cline{10-10} 
Mask Fusion (MF) &  & 0.0626 & 0.2250 & 0.0750 & 0.2103 & 0.7011 & 1.0 & 0.5159 & 0.00\tabularnewline
\cline{1-1} \cline{3-10} \cline{4-10} \cline{5-10} \cline{6-10} \cline{7-10} \cline{8-10} \cline{9-10} \cline{10-10} 
Voxblox++ (Vpp) &  & 0.2024 & 0.4521 & 0.2595 & 0.2827 & 0.6597 & 1.0 & 0.4774 & 0.00\tabularnewline
\cline{1-1} \cline{3-10} \cline{4-10} \cline{5-10} \cline{6-10} \cline{7-10} \cline{8-10} \cline{9-10} \cline{10-10} 
Kimera Semantics (KS) &  & 0.4434 & 0.8167 & 0.6037 & 0.7461 & 0.7718 & 1.0 & 0.6202 & 1.00\tabularnewline
\hline 
Semantic Fusion (SF) & \multirow{4}{*}{miniroom:1 (mr1)} & 0.2272 & 0.5926 & 0.2407 & 0.5532 & 0.7235 & 1.0 & 0.5512 & 0.00\tabularnewline
\cline{1-1} \cline{3-10} \cline{4-10} \cline{5-10} \cline{6-10} \cline{7-10} \cline{8-10} \cline{9-10} \cline{10-10} 
Mask Fusion (MF) &  & 0.2401 & 0.3611 & 0.3333 & 0.5225 & 0.6345 & 1.0 & 0.4534 & 0.00\tabularnewline
\cline{1-1} \cline{3-10} \cline{4-10} \cline{5-10} \cline{6-10} \cline{7-10} \cline{8-10} \cline{9-10} \cline{10-10} 
Voxblox++ (Vpp) &  & 0.1951 & 0.4815 & 0.2222 & 0.5316 & 0.6645 & 1.0 & 0.4742 & 1.00\tabularnewline
\cline{1-1} \cline{3-10} \cline{4-10} \cline{5-10} \cline{6-10} \cline{7-10} \cline{8-10} \cline{9-10} \cline{10-10} 
Kimera Semantics (KS) &  & 0.3844 & 0.7407 & 0.5185 & 0.7607 & 0.7607 & 1.0 & 0.6108 & 1.00\tabularnewline
\hline 
Semantic Fusion (SF) & \multirow{4}{*}{miniroom:3 (mr3)} & 0.3717 & 0.7576 & 0.4848 & 0.4589 & 0.6555 & 1.0 & 0.4863 & 0.00\tabularnewline
\cline{1-1} \cline{3-10} \cline{4-10} \cline{5-10} \cline{6-10} \cline{7-10} \cline{8-10} \cline{9-10} \cline{10-10} 
Mask Fusion (MF) &  & 0.2852 & 0.6970 & 0.4192 & 0.3082 & 0.7266 & 1.0 & 0.5807 & 0.00\tabularnewline
\cline{1-1} \cline{3-10} \cline{4-10} \cline{5-10} \cline{6-10} \cline{7-10} \cline{8-10} \cline{9-10} \cline{10-10} 
Voxblox++ (Vpp) &  & 0.2296 & 0.6566 & 0.2475 & 0.2640 & 0.5029 & 1.0 & 0.3028 & 0.00\tabularnewline
\cline{1-1} \cline{3-10} \cline{4-10} \cline{5-10} \cline{6-10} \cline{7-10} \cline{8-10} \cline{9-10} \cline{10-10} 
Kimera Semantics (KS) &  & 0.4606 & 0.8182 & 0.6364 & 0.6563 & 0.7735 & 1.0 & 0.6313 & 0.00\tabularnewline
\hline 
\hline 
Semantic Fusion (SF) & \multirow{4}{*}{Average (Avg.)} & 0.2509 & 0.5756 & 0.3243 & 0.4785 & 0.6637 & 1.0 & 0.4830 & 0.25\tabularnewline
\cline{1-1} \cline{3-10} \cline{4-10} \cline{5-10} \cline{6-10} \cline{7-10} \cline{8-10} \cline{9-10} \cline{10-10} 
Mask Fusion (MF) &  & 0.1971 & 0.3876 & 0.2737 & 0.3601 & 0.7277 & 1.0 & 0.5693 & 0.00\tabularnewline
\cline{1-1} \cline{3-10} \cline{4-10} \cline{5-10} \cline{6-10} \cline{7-10} \cline{8-10} \cline{9-10} \cline{10-10} 
Voxblox++ (Vpp) &  & 0.1885 & 0.4937 & 0.2031 & 0.3973 & 0.6164 & 1.0 & 0.4261 & 0.50\tabularnewline
\cline{1-1} \cline{3-10} \cline{4-10} \cline{5-10} \cline{6-10} \cline{7-10} \cline{8-10} \cline{9-10} \cline{10-10} 
Kimera Semantics (KS) &  & 0.4172 & 0.7790 & 0.5633 & 0.7276 & 0.7633 & 1.0 & 0.6141 & 0.75\tabularnewline
\hline 
\end{tabular}
\par\end{centering}
\centering{}\caption{Using all ground-truth data obtained from \cite{talbot2020benchbot}
(\textit{Case I}).\label{tab:Using-all-ground-truth}}
\vspace{-5mm}
\end{table*}

\begin{figure*}
\centering{} \includegraphics[width=1\textwidth]{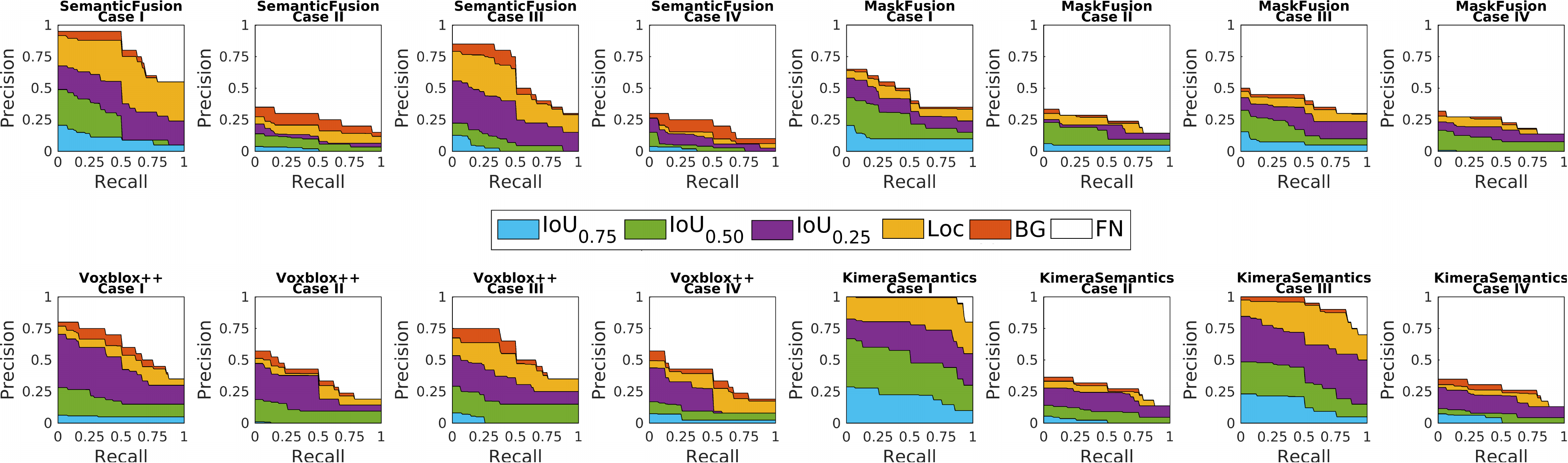}

\caption{Precision-Recall (PR) Curves. IoU$_{0.75}$, IoU$_{0.50}$, and IoU$_{0.25}$
are PR curves at IoU thresholds $0.75$, $0.50$, and $0.25$ respectively
over all four scenes. Loc is the PR curve at IoU threshold $0.10$
where localization errors are ignored while considering duplicate
detections. BG is the PR curve at IoU threshold $0.10$ after all
background false positives are removed. FN is the PR curve after all
errors are removed. (Note: the plots are inspired from \cite{cocoeval}.)
\label{fig:Precision-Recall-(PR)-Curves.}}
\vspace{-3.5mm}
\end{figure*}

We summarize the quantitative results of our experiments in Figs.~\ref{fig:scores}, ~\ref{fig:rscore}, and ~\ref{fig:Precision-Recall-(PR)-Curves.}.
Note that in Fig.~\ref{fig:rscore} we show the ratio of evaluation scores when compared to the best-case scenario (\textit{Case I}).
Our results show that in ideal conditions (\textit{Case I}),
methods that use class-aware segmentation like KS
and SF perform better than the methods that use instance-aware
segmentation like Vpp and MF because some objects
are not appropriately associated in a 3D map when seen from different
views. This is shown in detail in Fig.~\ref{fig:Object-Maps-}.
In Fig. \ref{fig:Object-Maps-}a, we see a chair and book, and in Fig. \ref{fig:Object-Maps-}d, we see a refrigerator, table and chair that do not have good data association when using MF and Vpp respectively. The same object
has been detected as a few fragmented cuboids. One of these fragments
may be taken as a true positive during the evaluation, while the rest
are considered false positives. In SF and KS, object instances are
extracted from the global semantic map after the mapping is completed,
unlike in MF and Vpp, where object instances in the map are built
incrementally. Hence, SF and KS seem to have a better data association
for the objects in the global map. Even when we use all ground-truth data
(\textit{Case I}), the resulting map is not perfect.
Fig. \ref{fig:Precision-Recall-(PR)-Curves.}
presents detailed analysis with different IoU thresholds and breakdown
of errors. Even when the localization error and background false positives are removed,
the mAP is not perfect. The main reason for this is that objects
or their parts are missed during the creation of the semantic 3D model
using the 2D object mask or segmentation and the depth image. In order
to filter out noise, these methods ignore objects a certain distance
away or of a certain size. Furthermore, some objects get missed or
only the parts of them get mapped because they have been seen from
limited views or very few times or both. In Fig. \ref{fig:scores}
mAP scores and in Fig. \ref{fig:Precision-Recall-(PR)-Curves.} area
under the PR curve decrease with an increase in the IoU threshold, which signifies
that only certain parts of the object get mapped. Since we have considered
that the object is fully confident to belong to a particular class
while building the object maps, the OMQ score solely depends upon
the spatial quality, i.e., IoU in 3D. When we look at Fig. \ref{fig:scores},
KS creates better maps than other methods based on OMQ and mAP metrics because the 3D global mesh produced by KS is very accurate \cite{Rosinol20icra-Kimera}.

When examining \textit{Case II}, which utilizes estimates of class- and instance-aware segmentation, we notice that the performance of all methods drops dramatically.
The mAP and OMQ performances in this scenario drop by up to 74.3\% and 71.3\% respectively under this single condition.
The performance drop is higher for SF and KS than
MF and Vpp, which we ascribe to the fact that Mask R-CNN produces more accurate object-wise segmentation masks than the class-wise segmentation masks of DilatedNet. 
This dramatic decrease in performance shows that accurate segmentation of classes/objects is a major limitation in attaining accurate semantic masks when using the current state-of-the-art semantic SLAM systems.

If instead we use ground-truth segmentation and instead estimate robot pose within the Semantic SLAM system (\textit{Case III}), we notice less of a reduction in semantic mapping performance. There is still a decrease in performance which can be attributed to the trajectory errors 
that we present in Table \ref{tab:RMSE-ATE-and}.
When comparing the orange and yellow bars in Fig.~\ref{fig:rscore} (\textit{Case II} and \textit{Case III} respectively), it is clear that segmentation error contributes more to low scores than the pose estimation error. The  mAP and  OMQ  performances  in  this  scenario  drop  by  up  to 41.7\%  and  34.7\%  respectively.

\begin{table}
\vspace{-1mm}
\begin{tabular}{|c|c|c|c|c|}
\hline 
\multirow{2}{*}{\textbf{Scene}} & \multirow{1}{*}{\textbf{Trajectory }} & \multirow{1}{*}{\textbf{ATE }} & \multirow{1}{*}{\textbf{RPE }} & \multirow{1}{*}{\textbf{RPE }}\tabularnewline
 & \textbf{Length (m)} & \textbf{(m)} & \textbf{Translation (m)} & \textbf{Rotation (deg)}\tabularnewline
\hline 
ap1 & 16.1973 & 0.0671 & 0.0056 & 0.6464\tabularnewline
\hline 
ap3 & 19.6031 & 0.0123 & 0.0051 & 0.1961\tabularnewline
\hline 
mr1 & 3.5404 & 0.0540 & 0.0185 & 1.3724\tabularnewline
\hline 
mr3 & 2.1417 & 0.0395 & 0.0109 & 0.3698\tabularnewline
\hline 
Avg. & 10.3706 & 0.0432 & 0.0100 & 0.6462\tabularnewline
\hline 
\end{tabular}

\caption{RMSE ATE and RPE error between ground truth trajectory and estimated
trajectory obtained from \cite{whelan2016elasticfusion}.\label{tab:RMSE-ATE-and}}

\vspace{-10mm}
\end{table}

\begin{table*}
\begin{tabular}{|c|c|c|c|c|c|c|c|c|c|c|c|c|c|}
\hline 
\multirow{2}{*}{\textbf{Method}} & \multirow{2}{*}{\textbf{Scene}} & \multirow{2}{*}{\textbf{$\mbox{\textbf{mAP}}_{\boldsymbol{\mbox{3D}}}$}} & \multirow{2}{*}{$\mbox{\textbf{mAP}}_{\boldsymbol{\mbox{3D}}}^{\mbox{\textbf{25}}}$} & \multirow{2}{*}{$\mbox{\textbf{mAP}}_{\boldsymbol{\mbox{3D}}}^{\mbox{\textbf{50}}}$} & \multirow{2}{*}{$\mbox{\textbf{OMQ}}$} & \multirow{2}{*}{\textbf{$\boldsymbol{\mbox{mPOQ}}$}} & \multirow{2}{*}{\textbf{$\boldsymbol{\mbox{mLQ}}$}} & \multirow{2}{*}{\textbf{$\boldsymbol{\mbox{mSQ}}$}} & \multirow{2}{*}{\textbf{$\boldsymbol{\mbox{mFPQ}}$}} & \multirow{2}{*}{\textbf{\textit{rmAP}}} & \multirow{2}{*}{\textbf{\textit{rAP25}}} & \multirow{2}{*}{\textbf{\textit{rAP50}}} & \multirow{2}{*}{\textbf{\textit{rOMQ}}}\tabularnewline
 &  &  &  &  &  &  &  &  &  &  &  &  & \tabularnewline
\hline 
SF & \multirow{4}{*}{ap1} & 0.0644 & 0.1346 & 0.0788 & 0.1330 & 0.6652 & 1.0 & 0.4649 & 0.00 & 0.2439 & 0.2431 & 0.2089 & 0.2441\tabularnewline
\cline{1-1} \cline{3-14} \cline{4-14} \cline{5-14} \cline{6-14} \cline{7-14} \cline{8-14} \cline{9-14} \cline{10-14} \cline{11-14} \cline{12-14} \cline{13-14} \cline{14-14} 
MF &  & 0.1292 & 0.2089 & 0.1768 & 0.2430 & 0.8263 & 1.0 & 0.6907 & 0.00 & 0.6446 & 0.7821 & 0.6618 & 0.6086\tabularnewline
\cline{1-1} \cline{3-14} \cline{4-14} \cline{5-14} \cline{6-14} \cline{7-14} \cline{8-14} \cline{9-14} \cline{10-14} \cline{11-14} \cline{12-14} \cline{13-14} \cline{14-14} 
Vpp &  & 0.0920 & 0.3083 & 0.0786 & 0.2421 & 0.6572 & 1.0 & 0.5161 & 0.00 & 0.7256 & 0.8014 & 0.9462 & 0.4740\tabularnewline
\cline{1-1} \cline{3-14} \cline{4-14} \cline{5-14} \cline{6-14} \cline{7-14} \cline{8-14} \cline{9-14} \cline{10-14} \cline{11-14} \cline{12-14} \cline{13-14} \cline{14-14} 
KS &  & 0.0618 & 0.1543 & 0.0971 & 0.2576 & 0.6583 & 1.0 & 0.4739 & 0.00 & 0.1626 & 0.2084 & 0.1964 & 0.3447\tabularnewline
\hline 
SF & \multirow{4}{*}{ap3} & 0.0475 & 0.1472 & 0.0375 & 0.1439 & 0.5515 & 1.0 & 0.3558 & 0.00 & 0.3381 & 0.3693 & 0.1929 & 0.4033\tabularnewline
\cline{1-1} \cline{3-14} \cline{4-14} \cline{5-14} \cline{6-14} \cline{7-14} \cline{8-14} \cline{9-14} \cline{10-14} \cline{11-14} \cline{12-14} \cline{13-14} \cline{14-14} 
MF &  & 0.0260 & 0.0556 & 0.0382 & 0.0942 & 0.5518 & 1.0 & 0.3824 & 0.00 & 0.4157 & 0.2469 & 0.5093 & 0.4479\tabularnewline
\cline{1-1} \cline{3-14} \cline{4-14} \cline{5-14} \cline{6-14} \cline{7-14} \cline{8-14} \cline{9-14} \cline{10-14} \cline{11-14} \cline{12-14} \cline{13-14} \cline{14-14} 
Vpp &  & 0.0634 & 0.2181 & 0.0549 & 0.1356 & 0.5086 & 1.0 & 0.3206 & 0.00 & 0.3132 & 0.4824 & 0.2114 & 0.4797\tabularnewline
\cline{1-1} \cline{3-14} \cline{4-14} \cline{5-14} \cline{6-14} \cline{7-14} \cline{8-14} \cline{9-14} \cline{10-14} \cline{11-14} \cline{12-14} \cline{13-14} \cline{14-14} 
KS &  & 0.0530 & 0.1806 & 0.0417 & 0.1949 & 0.6146 & 1.0 & 0.4204 & 0.00 & 0.1195 & 0.2211 & 0.0690 & 0.2612\tabularnewline
\hline 
SF & \multirow{4}{*}{mr1} & 0.0860 & 0.1975 & 0.1975 & 0.1405 & 0.5018 & 1.0 & 0.3399 & 0.00 & 0.3786 & 0.3333 & 0.8205 & 0.2540\tabularnewline
\cline{1-1} \cline{3-14} \cline{4-14} \cline{5-14} \cline{6-14} \cline{7-14} \cline{8-14} \cline{9-14} \cline{10-14} \cline{11-14} \cline{12-14} \cline{13-14} \cline{14-14} 
MF &  & 0.1580 & 0.2593 & 0.2222 & 0.1575 & 0.7089 & 1.0 & 0.5297 & 0.00 & 0.6581 & 0.7179 & 0.6667 & 0.3015\tabularnewline
\cline{1-1} \cline{3-14} \cline{4-14} \cline{5-14} \cline{6-14} \cline{7-14} \cline{8-14} \cline{9-14} \cline{10-14} \cline{11-14} \cline{12-14} \cline{13-14} \cline{14-14} 
Vpp &  & 0.0993 & 0.2012 & 0.2012 & 0.2110 & 0.5803 & 1.0 & 0.3871 & 0.00 & 0.5089 & 0.4179 & 0.9056 & 0.3970\tabularnewline
\cline{1-1} \cline{3-14} \cline{4-14} \cline{5-14} \cline{6-14} \cline{7-14} \cline{8-14} \cline{9-14} \cline{10-14} \cline{11-14} \cline{12-14} \cline{13-14} \cline{14-14} 
KS &  & 0.1654 & 0.4259 & 0.2593 & 0.2856 & 0.6784 & 1.0 & 0.4809 & 0.00 & 0.4304 & 0.5750 & 0.5000 & 0.3755\tabularnewline
\hline 
SF & \multirow{4}{*}{mr3} & 0.0970 & 0.2424 & 0.1515 & 0.1324 & 0.6288 & 1.0 & 0.4543 & 0.00 & 0.2609 & 0.3200 & 0.3125 & 0.2885\tabularnewline
\cline{1-1} \cline{3-14} \cline{4-14} \cline{5-14} \cline{6-14} \cline{7-14} \cline{8-14} \cline{9-14} \cline{10-14} \cline{11-14} \cline{12-14} \cline{13-14} \cline{14-14} 
MF &  & 0.1111 & 0.2121 & 0.1212 & 0.1452 & 0.6242 & 1.0 & 0.4479 & 0.00 & 0.3896 & 0.3043 & 0.2892 & 0.4709\tabularnewline
\cline{1-1} \cline{3-14} \cline{4-14} \cline{5-14} \cline{6-14} \cline{7-14} \cline{8-14} \cline{9-14} \cline{10-14} \cline{11-14} \cline{12-14} \cline{13-14} \cline{14-14} 
Vpp &  & 0.1343 & 0.3485 & 0.1667 & 0.1554 & 0.4557 & 1.0 & 0.2586 & 0.00 & 0.5850 & 0.5308 & 0.6735 & 0.5885\tabularnewline
\cline{1-1} \cline{3-14} \cline{4-14} \cline{5-14} \cline{6-14} \cline{7-14} \cline{8-14} \cline{9-14} \cline{10-14} \cline{11-14} \cline{12-14} \cline{13-14} \cline{14-14} 
KS &  & 0.1483 & 0.3712 & 0.1919 & 0.2165 & 0.6341 & 1.0 & 0.4492 & 0.00 & 0.3220 & 0.4537 & 0.3016 & 0.3299\tabularnewline
\hline 
\hline 
SF & \multirow{4}{*}{Avg.} & 0.0737 & 0.1805 & 0.1163 & 0.1375 & 0.5868 & 1.0 & 0.4037 & 0.00 & 0.2939 & 0.3135 & 0.3587 & 0.2873\tabularnewline
\cline{1-1} \cline{3-14} \cline{4-14} \cline{5-14} \cline{6-14} \cline{7-14} \cline{8-14} \cline{9-14} \cline{10-14} \cline{11-14} \cline{12-14} \cline{13-14} \cline{14-14} 
MF &  & 0.1061 & 0.1840 & 0.1396 & 0.1600 & 0.6778 & 1.0 & 0.5127 & 0.00 & 0.5383 & 0.4747 & 0.5101 & 0.4443\tabularnewline
\cline{1-1} \cline{3-14} \cline{4-14} \cline{5-14} \cline{6-14} \cline{7-14} \cline{8-14} \cline{9-14} \cline{10-14} \cline{11-14} \cline{12-14} \cline{13-14} \cline{14-14} 
Vpp &  & 0.0972 & 0.2690 & 0.1253 & 0.1860 & 0.5504 & 1.0 & 0.3706 & 0.00 & 0.5160 & 0.5449 & 0.6172 & 0.4682\tabularnewline
\cline{1-1} \cline{3-14} \cline{4-14} \cline{5-14} \cline{6-14} \cline{7-14} \cline{8-14} \cline{9-14} \cline{10-14} \cline{11-14} \cline{12-14} \cline{13-14} \cline{14-14} 
KS &  & 0.1071 & 0.2830 & 0.1475 & 0.2386 & 0.6463 & 1.0 & 0.4561 & 0.00 & 0.2568 & 0.3633 & 0.2618 & 0.3280\tabularnewline
\hline 
\end{tabular}\caption{Using ground-truth pose from \cite{talbot2020benchbot} and estimated
segmentation \cite{zhou2016semantic} or object masks \cite{he2017mask}
(\textit{Case II}).\label{tab:Using-ground-truth-pose}}

\centering{}\vspace{-7mm}
\end{table*}
\begin{table*}
\begin{tabular}{|c|c|c|c|c|c|c|c|c|c|c|c|c|c|}
\hline 
\multirow{2}{*}{\textbf{Method}} & \multirow{2}{*}{\textbf{Scene}} & \multirow{2}{*}{\textbf{$\mbox{\textbf{mAP}}_{\boldsymbol{\mbox{3D}}}$}} & \multirow{2}{*}{$\mbox{\textbf{mAP}}_{\boldsymbol{\mbox{3D}}}^{\mbox{\textbf{25}}}$} & \multirow{2}{*}{$\mbox{\textbf{mAP}}_{\boldsymbol{\mbox{3D}}}^{\mbox{\textbf{50}}}$} & \multirow{2}{*}{$\mbox{\textbf{OMQ}}$} & \multirow{2}{*}{\textbf{$\boldsymbol{\mbox{mPOQ}}$}} & \multirow{2}{*}{\textbf{$\boldsymbol{\mbox{mLQ}}$}} & \multirow{2}{*}{\textbf{$\boldsymbol{\mbox{mSQ}}$}} & \multirow{2}{*}{\textbf{$\boldsymbol{\mbox{mFPQ}}$}} & \multirow{2}{*}{\textbf{\textit{rmAP}}} & \multirow{2}{*}{\textbf{\textit{rAP25}}} & \multirow{2}{*}{\textbf{\textit{rAP50}}} & \multirow{2}{*}{\textbf{\textit{rOMQ}}}\tabularnewline
 &  &  &  &  &  &  &  &  &  &  &  &  & \tabularnewline
\hline 
SF & \multirow{4}{*}{ap1} & 0.1586 & 0.4452 & 0.1417 & 0.1672 & 0.5853 & 1.0 & 0.3536 & 0.00 & 0.6004 & 0.8040 & 0.3756 & 0.3068\tabularnewline
\cline{1-1} \cline{3-14} \cline{4-14} \cline{5-14} \cline{6-14} \cline{7-14} \cline{8-14} \cline{9-14} \cline{10-14} \cline{11-14} \cline{12-14} \cline{13-14} \cline{14-14} 
MF &  & 0.1354 & 0.2420 & 0.1705 & 0.2632 & 0.7020 & 1.0 & 0.5000 & 0.00 & 0.6755 & 0.9057 & 0.6384 & 0.6592\tabularnewline
\cline{1-1} \cline{3-14} \cline{4-14} \cline{5-14} \cline{6-14} \cline{7-14} \cline{8-14} \cline{9-14} \cline{10-14} \cline{11-14} \cline{12-14} \cline{13-14} \cline{14-14} 
Vpp &  & 0.0490 & 0.2655 & 0.0274 & 0.2219 & 0.5446 & 1.0 & 0.3388 & 0.00 & 0.3869 & 0.6900 & 0.3297 & 0.4343\tabularnewline
\cline{1-1} \cline{3-14} \cline{4-14} \cline{5-14} \cline{6-14} \cline{7-14} \cline{8-14} \cline{9-14} \cline{10-14} \cline{11-14} \cline{12-14} \cline{13-14} \cline{14-14} 
KS &  & 0.2683 & 0.6619 & 0.3402 & 0.4398 & 0.6597 & 1.0 & 0.4709 & 0.00 & 0.7055 & 0.8942 & 0.6876 & 0.5885\tabularnewline
\hline 
SF & \multirow{4}{*}{ap3} & 0.0958 & 0.1890 & 0.1500 & 0.2637 & 0.5670 & 1.0 & 0.3768 & 0.00 & 0.6812 & 0.4742 & 0.7714 & 0.7393\tabularnewline
\cline{1-1} \cline{3-14} \cline{4-14} \cline{5-14} \cline{6-14} \cline{7-14} \cline{8-14} \cline{9-14} \cline{10-14} \cline{11-14} \cline{12-14} \cline{13-14} \cline{14-14} 
MF &  & 0.0429 & 0.2155 & 0.0472 & 0.1881 & 0.6268 & 1.0 & 0.4369 & 0.00 & 0.6862 & 0.9577 & 0.6296 & 0.8941\tabularnewline
\cline{1-1} \cline{3-14} \cline{4-14} \cline{5-14} \cline{6-14} \cline{7-14} \cline{8-14} \cline{9-14} \cline{10-14} \cline{11-14} \cline{12-14} \cline{13-14} \cline{14-14} 
Vpp &  & 0.1547 & 0.3535 & 0.1868 & 0.2780 & 0.6115 & 1.0 & 0.4344 & 0.00 & 0.7645 & 0.7819 & 0.7198 & 0.9831\tabularnewline
\cline{1-1} \cline{3-14} \cline{4-14} \cline{5-14} \cline{6-14} \cline{7-14} \cline{8-14} \cline{9-14} \cline{10-14} \cline{11-14} \cline{12-14} \cline{13-14} \cline{14-14} 
KS &  & 0.3324 & 0.6500 & 0.3764 & 0.6676 & 0.7153 & 1.0 & 0.5545 & 1.00 & 0.7496 & 0.7959 & 0.6235 & 0.8948\tabularnewline
\hline 
SF & \multirow{4}{*}{mr1} & 0.0383 & 0.2407 & 0.0370 & 0.4416 & 0.6992 & 1.0 & 0.5231 & 0.00 & 0.1685 & 0.4062 & 0.1538 & 0.7982\tabularnewline
\cline{1-1} \cline{3-14} \cline{4-14} \cline{5-14} \cline{6-14} \cline{7-14} \cline{8-14} \cline{9-14} \cline{10-14} \cline{11-14} \cline{12-14} \cline{13-14} \cline{14-14} 
MF &  & 0.1043 & 0.2870 & 0.2222 & 0.2874 & 0.8623 & 1.0 & 0.7448 & 1.00 & 0.4344 & 0.7949 & 0.6667 & 0.5501\tabularnewline
\cline{1-1} \cline{3-14} \cline{4-14} \cline{5-14} \cline{6-14} \cline{7-14} \cline{8-14} \cline{9-14} \cline{10-14} \cline{11-14} \cline{12-14} \cline{13-14} \cline{14-14} 
Vpp &  & 0.1280 & 0.3025 & 0.1852 & 0.3840 & 0.5761 & 1.0 & 0.3878 & 0.00 & 0.6561 & 0.6282 & 0.8333 & 0.7224\tabularnewline
\cline{1-1} \cline{3-14} \cline{4-14} \cline{5-14} \cline{6-14} \cline{7-14} \cline{8-14} \cline{9-14} \cline{10-14} \cline{11-14} \cline{12-14} \cline{13-14} \cline{14-14} 
KS &  & 0.2490 & 0.4630 & 0.3333 & 0.6880 & 0.7339 & 1.0 & 0.5632 & 0.00 & 0.6478 & 0.6250 & 0.6429 & 0.9044\tabularnewline
\hline 
SF & \multirow{4}{*}{mr3} & 0.2929 & 0.6667 & 0.2727 & 0.4148 & 0.6221 & 1.0 & 0.4389 & 0.00 & 0.7880 & 0.8800 & 0.5625 & 0.9038\tabularnewline
\cline{1-1} \cline{3-14} \cline{4-14} \cline{5-14} \cline{6-14} \cline{7-14} \cline{8-14} \cline{9-14} \cline{10-14} \cline{11-14} \cline{12-14} \cline{13-14} \cline{14-14} 
MF &  & 0.2263 & 0.3333 & 0.3333 & 0.2513 & 0.6462 & 1.0 & 0.4747 & 0.00 & 0.7934 & 0.4783 & 0.7952 & 0.8153\tabularnewline
\cline{1-1} \cline{3-14} \cline{4-14} \cline{5-14} \cline{6-14} \cline{7-14} \cline{8-14} \cline{9-14} \cline{10-14} \cline{11-14} \cline{12-14} \cline{13-14} \cline{14-14} 
Vpp &  & 0.1946 & 0.5354 & 0.1919 & 0.1539 & 0.4345 & 1.0 & 0.2402 & 0.00 & 0.8475 & 0.8154 & 0.7755 & 0.5829\tabularnewline
\cline{1-1} \cline{3-14} \cline{4-14} \cline{5-14} \cline{6-14} \cline{7-14} \cline{8-14} \cline{9-14} \cline{10-14} \cline{11-14} \cline{12-14} \cline{13-14} \cline{14-14} 
KS &  & 0.3655 & 0.7955 & 0.5227 & 0.5592 & 0.6590 & 1.0 & 0.4829 & 0.00 & 0.7935 & 0.9722 & 0.8214 & 0.8520\tabularnewline
\hline 
\hline 
SF & \multirow{4}{*}{Avg.} & 0.1464 & 0.3854 & 0.1504 & 0.3218 & 0.6184 & 1.0 & 0.4231 & 0.00 & 0.5835 & 0.6695 & 0.4636 & 0.6726\tabularnewline
\cline{1-1} \cline{3-14} \cline{4-14} \cline{5-14} \cline{6-14} \cline{7-14} \cline{8-14} \cline{9-14} \cline{10-14} \cline{11-14} \cline{12-14} \cline{13-14} \cline{14-14} 
MF &  & 0.1272 & 0.2695 & 0.1933 & 0.2475 & 0.7093 & 1.0 & 0.5391 & 0.25 & 0.6456 & 0.6953 & 0.7064 & 0.6873\tabularnewline
\cline{1-1} \cline{3-14} \cline{4-14} \cline{5-14} \cline{6-14} \cline{7-14} \cline{8-14} \cline{9-14} \cline{10-14} \cline{11-14} \cline{12-14} \cline{13-14} \cline{14-14} 
Vpp &  & 0.1316 & 0.3642 & 0.1478 & 0.2594 & 0.5417 & 1.0 & 0.3503 & 0.00 & 0.6983 & 0.7377 & 0.7280 & 0.6530\tabularnewline
\cline{1-1} \cline{3-14} \cline{4-14} \cline{5-14} \cline{6-14} \cline{7-14} \cline{8-14} \cline{9-14} \cline{10-14} \cline{11-14} \cline{12-14} \cline{13-14} \cline{14-14} 
KS &  & 0.3038 & 0.6426 & 0.3932 & 0.5887 & 0.6920 & 1.0 & 0.5179 & 0.25 & 0.7282 & 0.8249 & 0.6979 & 0.8090\tabularnewline
\hline 
\end{tabular}\caption{Using ground-truth segmentation or object masks from \cite{talbot2020benchbot}
and estimated pose from \cite{whelan2016elasticfusion} (\textit{Case
III}).\label{tab:Using-ground-truth-segmentation}}

\vspace{-5mm}
\end{table*}
\begin{table*}
\begin{tabular}{|c|c|c|c|c|c|c|c|c|c|c|c|c|c|}
\hline 
\multirow{2}{*}{\textbf{Method}} & \multirow{2}{*}{\textbf{Scene}} & \multirow{2}{*}{\textbf{$\mbox{\textbf{mAP}}_{\boldsymbol{\mbox{3D}}}$}} & \multirow{2}{*}{$\mbox{\textbf{mAP}}_{\boldsymbol{\mbox{3D}}}^{\mbox{\textbf{25}}}$} & \multirow{2}{*}{$\mbox{\textbf{mAP}}_{\boldsymbol{\mbox{3D}}}^{\mbox{\textbf{50}}}$} & \multirow{2}{*}{$\mbox{\textbf{OMQ}}$} & \multirow{2}{*}{\textbf{$\boldsymbol{\mbox{mPOQ}}$}} & \multirow{2}{*}{\textbf{$\boldsymbol{\mbox{mLQ}}$}} & \multirow{2}{*}{\textbf{$\boldsymbol{\mbox{mSQ}}$}} & \multirow{2}{*}{\textbf{$\boldsymbol{\mbox{mFPQ}}$}} & \multirow{2}{*}{\textbf{\textit{rmAP}}} & \multirow{2}{*}{\textbf{\textit{rAP25}}} & \multirow{2}{*}{\textbf{\textit{rAP50}}} & \multirow{2}{*}{\textbf{\textit{rOMQ}}}\tabularnewline
 &  &  &  &  &  &  &  &  &  &  &  &  & \tabularnewline
\hline 
SF & \multirow{4}{*}{ap1} & 0.0542 & 0.1429 & 0.0713 & 0.0891 & 0.5167 & 1.0 & 0.2720 & 0.00 & 0.2050 & 0.2580 & 0.1891 & 0.1635\tabularnewline
\cline{1-1} \cline{3-14} \cline{4-14} \cline{5-14} \cline{6-14} \cline{7-14} \cline{8-14} \cline{9-14} \cline{10-14} \cline{11-14} \cline{12-14} \cline{13-14} \cline{14-14} 
MF &  & 0.0776 & 0.2000 & 0.0964 & 0.1928 & 0.6939 & 1.0 & 0.5326 & 0.00 & 0.3874 & 0.7487 & 0.3610 & 0.4827\tabularnewline
\cline{1-1} \cline{3-14} \cline{4-14} \cline{5-14} \cline{6-14} \cline{7-14} \cline{8-14} \cline{9-14} \cline{10-14} \cline{11-14} \cline{12-14} \cline{13-14} \cline{14-14} 
Vpp &  & 0.0824 & 0.2430 & 0.0762 & 0.0980 & 0.4760 & 1.0 & 0.2736 & 0.00 & 0.6500 & 0.6316 & 0.9176 & 0.1918\tabularnewline
\cline{1-1} \cline{3-14} \cline{4-14} \cline{5-14} \cline{6-14} \cline{7-14} \cline{8-14} \cline{9-14} \cline{10-14} \cline{11-14} \cline{12-14} \cline{13-14} \cline{14-14} 
KS &  & 0.0355 & 0.1348 & 0.0515 & 0.2090 & 0.6531 & 1.0 & 0.4465 & 0.00 & 0.0934 & 0.1821 & 0.1041 & 0.2797\tabularnewline
\hline 
SF & \multirow{4}{*}{ap3} & 0.0479 & 0.1233 & 0.0521 & 0.1243 & 0.5386 & 1.0 & 0.3385 & 0.00 & 0.3407 & 0.3094 & 0.2679 & 0.3484\tabularnewline
\cline{1-1} \cline{3-14} \cline{4-14} \cline{5-14} \cline{6-14} \cline{7-14} \cline{8-14} \cline{9-14} \cline{10-14} \cline{11-14} \cline{12-14} \cline{13-14} \cline{14-14} 
MF &  & 0.0239 & 0.0556 & 0.0330 & 0.0928 & 0.5434 & 1.0 & 0.3737 & 0.00 & 0.3817 & 0.2469 & 0.4398 & 0.4411\tabularnewline
\cline{1-1} \cline{3-14} \cline{4-14} \cline{5-14} \cline{6-14} \cline{7-14} \cline{8-14} \cline{9-14} \cline{10-14} \cline{11-14} \cline{12-14} \cline{13-14} \cline{14-14} 
Vpp &  & 0.0582 & 0.1486 & 0.0562 & 0.1548 & 0.4764 & 1.0 & 0.3030 & 0.00 & 0.2875 & 0.3287 & 0.2168 & 0.5476\tabularnewline
\cline{1-1} \cline{3-14} \cline{4-14} \cline{5-14} \cline{6-14} \cline{7-14} \cline{8-14} \cline{9-14} \cline{10-14} \cline{11-14} \cline{12-14} \cline{13-14} \cline{14-14} 
KS &  & 0.0475 & 0.1847 & 0.0417 & 0.1884 & 0.6230 & 1.0 & 0.4309 & 0.00 & 0.1071 & 0.2262 & 0.0690 & 0.2525\tabularnewline
\hline 
SF & \multirow{4}{*}{mr1} & 0.0173 & 0.1481 & 0.0000 & 0.1397 & 0.5121 & 1.0 & 0.3133 & 0.00 & 0.0761 & 0.2500 & 0.0000 & 0.2525\tabularnewline
\cline{1-1} \cline{3-14} \cline{4-14} \cline{5-14} \cline{6-14} \cline{7-14} \cline{8-14} \cline{9-14} \cline{10-14} \cline{11-14} \cline{12-14} \cline{13-14} \cline{14-14} 
MF &  & 0.1105 & 0.2130 & 0.1667 & 0.1574 & 0.7083 & 1.0 & 0.5318 & 0.00 & 0.4602 & 0.5897 & 0.5000 & 0.3013\tabularnewline
\cline{1-1} \cline{3-14} \cline{4-14} \cline{5-14} \cline{6-14} \cline{7-14} \cline{8-14} \cline{9-14} \cline{10-14} \cline{11-14} \cline{12-14} \cline{13-14} \cline{14-14} 
Vpp &  & 0.0794 & 0.1493 & 0.1111 & 0.2147 & 0.6440 & 1.0 & 0.4486 & 0.00 & 0.4068 & 0.3101 & 0.5000 & 0.4038\tabularnewline
\cline{1-1} \cline{3-14} \cline{4-14} \cline{5-14} \cline{6-14} \cline{7-14} \cline{8-14} \cline{9-14} \cline{10-14} \cline{11-14} \cline{12-14} \cline{13-14} \cline{14-14} 
KS &  & 0.1321 & 0.3642 & 0.1790 & 0.2723 & 0.6467 & 1.0 & 0.4333 & 0.00 & 0.3437 & 0.4917 & 0.3452 & 0.3580\tabularnewline
\hline 
SF & \multirow{4}{*}{mr3} & 0.0758 & 0.2424 & 0.1061 & 0.1275 & 0.5421 & 1.0 & 0.3686 & 0.00 & 0.2038 & 0.3200 & 0.2187 & 0.2779\tabularnewline
\cline{1-1} \cline{3-14} \cline{4-14} \cline{5-14} \cline{6-14} \cline{7-14} \cline{8-14} \cline{9-14} \cline{10-14} \cline{11-14} \cline{12-14} \cline{13-14} \cline{14-14} 
MF &  & 0.1152 & 0.2121 & 0.2121 & 0.1259 & 0.5413 & 1.0 & 0.3399 & 0.00 & 0.4038 & 0.3043 & 0.5060 & 0.4084\tabularnewline
\cline{1-1} \cline{3-14} \cline{4-14} \cline{5-14} \cline{6-14} \cline{7-14} \cline{8-14} \cline{9-14} \cline{10-14} \cline{11-14} \cline{12-14} \cline{13-14} \cline{14-14} 
Vpp &  & 0.1394 & 0.3636 & 0.1515 & 0.1004 & 0.4866 & 1.0 & 0.2792 & 0.00 & 0.6070 & 0.5538 & 0.6122 & 0.3803\tabularnewline
\cline{1-1} \cline{3-14} \cline{4-14} \cline{5-14} \cline{6-14} \cline{7-14} \cline{8-14} \cline{9-14} \cline{10-14} \cline{11-14} \cline{12-14} \cline{13-14} \cline{14-14} 
KS &  & 0.1019 & 0.3258 & 0.1465 & 0.1900 & 0.5565 & 1.0 & 0.3563 & 0.00 & 0.2211 & 0.3981 & 0.2302 & 0.2895\tabularnewline
\hline 
\hline 
SF & \multirow{4}{*}{Avg.} & 0.0488 & 0.1642 & 0.0574 & 0.1202 & 0.5274 & 1.0 & 0.3231 & 0.00 & 0.1944 & 0.2852 & 0.1769 & 0.2511\tabularnewline
\cline{1-1} \cline{3-14} \cline{4-14} \cline{5-14} \cline{6-14} \cline{7-14} \cline{8-14} \cline{9-14} \cline{10-14} \cline{11-14} \cline{12-14} \cline{13-14} \cline{14-14} 
MF &  & 0.0818 & 0.1702 & 0.1271 & 0.1422 & 0.6217 & 1.0 & 0.4445 & 0.00 & 0.4150 & 0.4391 & 0.4643 & 0.3949\tabularnewline
\cline{1-1} \cline{3-14} \cline{4-14} \cline{5-14} \cline{6-14} \cline{7-14} \cline{8-14} \cline{9-14} \cline{10-14} \cline{11-14} \cline{12-14} \cline{13-14} \cline{14-14} 
Vpp &  & 0.0898 & 0.2261 & 0.0988 & 0.1420 & 0.5208 & 1.0 & 0.3261 & 0.00 & 0.4767 & 0.4580 & 0.4864 & 0.3574\tabularnewline
\cline{1-1} \cline{3-14} \cline{4-14} \cline{5-14} \cline{6-14} \cline{7-14} \cline{8-14} \cline{9-14} \cline{10-14} \cline{11-14} \cline{12-14} \cline{13-14} \cline{14-14} 
KS &  & 0.0792 & 0.2524 & 0.1047 & 0.2149 & 0.6199 & 1.0 & 0.4167 & 0.00 & 0.1899 & 0.3240 & 0.1858 & 0.2954\tabularnewline
\hline 
\end{tabular}\caption{Using estimated pose \cite{whelan2016elasticfusion} and estimated
segmentation \cite{zhou2016semantic} or object masks \cite{he2017mask}
(\textit{Case I}V).\label{tab:Using-estimated-pose}}

\vspace{-7mm}
\end{table*}

When we consider the case where both segmentation and pose are estimated (\textit{Case IV}), shown in green (Fig.~\ref{fig:rscore}), we see the drop in performance does not far exceed that of \textit{Case II} in orange.
The average performance drops to $0.401$, $0.664$, and $0.319$ 
for Cases \textit{II}, \textit{III} and \textit{IV} respectively compared
to \textit{Case I} based on mAP. The average performance drops to $0.382$, $0.705$, and $0.325$ respectively based on OMQ.
\textit{Case IV} is the scenario most equivalent to the real robotic scenario but in a simulated environment. 
From our results, we conclude that while both segmentation accuracy and pose estimation affect the accuracy of semantic maps in semantic SLAM systems, the impact of segmentation accuracy far exceeds that of pose estimation and should be the focus of study within this field of research.

\section{CONCLUSIONS}
In this paper, we evaluated the impact of semantic segmentation and pose estimation on the semantic maps produced by class- and instance-aware dense semantic SLAM systems.
We have presented this evaluation methodology in an open-format through the BenchBot framework, and encourage future research to evaluate performance using these tools.
By using simulated environments to provide semantic SLAM algorithms with ground-truth pose and/or semantic segmentation data, we analyzed each component individually for different systems.
In comparing the ground-truth cuboid maps with the cuboid maps extracted from the semantic SLAM systems, we have demonstrated that the largest source of error is semantic segmentation.
Semantic segmentation was shown to drop mAP and OMQ performance by up to 74.3\% and 71.3\% respectively.
It is also shown that better data association results in more accurate instance-aware maps. We also observe that providing  perfect  segmentation,  depth  and  pose  data alone does not guarantee perfect semantic mapping if we do not get all the objects or all the parts of the objects in the view. This aspect of SLAM, how to explore the environment to map all the objects is also important to improve the quality of semantic map.
We hope that these observations will inspire or motivate further research in semantic SLAM and semantic mapping of the environment.




\bibliographystyle{ieeetr}
\bibliography{ref}

\end{document}